\pdfoutput=1

\documentclass[11pt]{article}

\usepackage{acl}

\usepackage{times}
\usepackage{latexsym}
\usepackage{amsmath}

\usepackage[T1]{fontenc}

\usepackage[utf8]{inputenc}

\usepackage{microtype}
\usepackage{amsfonts}
\usepackage{dsfont}
\usepackage{graphicx}
\usepackage{booktabs}
\usepackage{tabularx}
\usepackage{bm}

\title{Unveiling the Achilles' Heel of NLG Evaluators: \\A Unified Adversarial Framework Driven by Large Language Models}

\author{Yiming Chen$^{\dag}$ \quad Chen Zhang$^{\dag\thanks{\quad Corresponding author.}}$ \quad Danqing Luo$^{\dag}$ \quad Luis Fernando D'Haro$^{\star}$ \\ \textbf{Robby T. Tan}$^{\ddag, \dag}$ \quad \textbf{Haizhou Li}$^{\natural, \dag, \S}$ \\
        $^\dag$National University of Singapore \quad 
        $^{\natural}$The Chinese University of Hong Kong, Shenzhen \\
        $^{\star}$Universidad Politécnica de Madrid \quad
        $^{\ddag}$ASUS Intelligent Cloud Services\quad
        $^{\S}$Kriston AI Lab\\
        \tt \{yiming.chen,chen\_zhang\}@u.nus.edu  \\ \tt danqing@nus.edu.sg\quad luisfernando.dharo@upm.es  \\
   \tt  robby\_tan@asus.com\quad haizhouli@cuhk.edu.cn \\
}

\begin{document}
\maketitle

\begin{abstract}
The automatic evaluation of natural language generation (NLG) systems presents a long-lasting challenge. Recent studies have highlighted various neural metrics that align well with human evaluations. Yet, the robustness of these evaluators against adversarial perturbations remains largely under-explored due to the unique challenges in obtaining adversarial data for different NLG evaluation tasks. 
To address the problem, we introduce AdvEval, a novel black-box adversarial framework against NLG evaluators. 
AdvEval is specially tailored to generate data that yield strong disagreements between human and victim evaluators. 
Specifically, inspired by the recent success of large language models (LLMs) in text generation and evaluation, we adopt strong LLMs as both the data generator and gold evaluator. Adversarial data are automatically optimized with feedback from the gold and victim evaluator. We conduct experiments on 12 victim evaluators and 11 NLG datasets, spanning tasks including dialogue, summarization, and question evaluation. The results show that AdvEval can lead to significant performance degradation of various victim metrics, thereby validating its efficacy.\footnote{Code is available at \url{github.com/MatthewCYM/AdvEval}.}

\end{abstract}
\section{Introduction}
Recent advancements in generative language models, especially advanced large language models (LLMs), have led to their widespread application in creating human-like content across various domains. Concurrently, numerous automatic text evaluation metrics have been developed, aiming to identify unhelpful responses and understand the model's capacity. These metrics have exhibited good correlations with human judgment across diverse natural language generation (NLG) tasks, e.g., dialogue generation~\citep{zhang-etal-2021-dynaeval}, summarization~\citep{zhong-etal-2022-towards,yuan2021bartscore}, question generation~\citep{mohammadshahi-etal-2023-rquge,wang-etal-2022-qrelscore}. Despite their strong performance, the robustness of these metrics against adversarial attacks or misleading inputs remains critically under-explored. Therefore, in this paper, we aim to automatically identify "Achilles' Heel", critical weaknesses, within well-established automatic evaluation metrics.

Adversarial attack techniques are widely adopted to assess the robustness of neural models. Yet, their application in NLP has predominantly focused on general classification tasks, e.g., sentiment classification~\citep{Jin_Jin_Zhou_Szolovits_2020,DBLP:conf/ndss/LiJDLW19}. These attack techniques all build upon the label-preserving assumption that the gold label of input text will remain unchanged after adding bounded perturbation on the input text~\citep{wang-etal-2022-measure}. This assumption, however, does not hold on the NLG evaluation tasks, where even slight modifications can drastically alter, often negatively, a text's perceived quality, which directly correlates with the regression label in the evaluation tasks. 
For example, changing a single word "movie" to "dish" in the dialogue response "That's my favorite movie" will perfectly preserve the positive sentiment classification label but significantly alter the response relevance evaluation label, given that the dialogue context is about a movie.
Consequently, these traditional attacking techniques can easily obtain low-quality adversarial texts with high metric scores while struggling to produce high-quality adversarial texts that score low. Existing studies on NLG evaluators' robustness mainly rely on manual effort or heuristic perturbation rules to ensure the perturbed text accords with specified labels. These studies also fail to obtain high-quality adversarial texts with low metric scores and cannot scale up to generate large-scale diverse adversarial data~\citep{khalid-lee-2022-explaining, sai-etal-2021-perturbation}.

To overcome these limitations, we propose AdvEval, designed to obtain data that lead to two different types of disagreement between humans and victim evaluators: 1) text valued by humans but underrated by victim evaluators; 2) text underrated by humans but valued by victim evaluators. Inspired by recent works on using LLMs to iteratively improve their response based on current feedback~\citep{madaan2023self,yang2023large, chen2023teaching}, we adapt LLMs as the data generator to iteratively optimize the benign input text with the adversarial objective. As mentioned above, label preservation posts a unique challenge in attacking NLG tasks. Recognizing the recent trend of leveraging large language models (LLMs) as a substitute for human annotators to generate and annotate high-quality data~\cite{he2023annollm,ye-etal-2022-zerogen,ye-etal-2022-progen,meng2022generating}, we also utilize LLMs as the evaluator to give feedback during the optimization process, thereby imposing an implicit quality constraint on the generated text.

We evaluate AdvEval on 11 datasets covering three distinct NLG tasks against 12 automatic evaluation metrics that vary in their reliance on references, models, and LLMs. Experimental results demonstrate AdvEval's ability to generate high-quality adversarial data effectively. Through manual inspection, we also confirm the label preservation property of AdvEval. Further experiments on adversarial learning, ablation studies, and case studies convincingly validate the effectiveness and generalization ability of our proposed AdvEval framework.

\section{Background}
\subsection{NLG Evaluator}
The evaluation of AI-generated content is increasingly important recently in various domains~\citep{su2020blindly,9023056,jiang2023tigerscore}. Automatic NLG evaluators can be broadly divided into reference-based and reference-free metrics. Traditional reference-based metrics like BLEU~\citep{papineni-etal-2002-bleu} and ROUGE~\citep{lin-2004-rouge} are favored for their simplicity and versatility across NLG tasks. Yet, their inability to capture complex textual semantics has led to the development of newer reference-based metrics utilizing contextualized embeddings  for enhanced semantic understanding, such as BERTScore~\citep{Zhang*2020BERTScore:}, MoverScore~\citep{zhao-etal-2019-moverscore}, and BLEURT~\citep{sellam-etal-2020-bleurt}.

In open-ended tasks like dialogue generation, reference-based metrics often fall short in mirroring human judgment~\citep{liu-etal-2016-evaluate}. This has led to the development of trained reference-free metrics such as RUBER~\citep{Tao_Mou_Zhao_Yan_2018}, DynaEval~\citep{zhang-etal-2021-dynaeval}, UniEval~\citep{zhong-etal-2022-towards}, TRUE~\citep{honovich-etal-2022-true-evaluating}, and PoE~\citep{zhang2023poe}. These reference free metrics offer closer alignment with human evaluations but still don't reach the criteria for strong automatic evaluators\footnote{Defined as having a correlation over 0.8 with human evaluations.} and remain task-specific.

The emergence of LLMs marks a significant step towards developing more general and effective NLG evaluators. Several recent works~\cite{chen2023exploring,liu2023gpteval,lin-chen-2023-llm} examine the NLG evaluation capability of instruction-following LLMs via prompting and report strong correlations with human evaluation. Remarkably, GPT-4~\citep{openai2023gpt4} is found to exhibit human-level judgment while being more scalable and less expensive than crowd-sourced workers~\citep{zheng2023judging,gilardi2023chatgpt,chiang-lee-2023-large}. Inspired by these findings, our AdvEval framework incorporates advanced proprietary LLMs like GPT-4 and PALM-2 to serve as proxies for human evaluators. 
Moreover, we conduct a comprehensive analysis of the robustness of metrics from the aforementioned types\footnote{Including reference-based and reference-free, trained and untrained, as well as BERT-based and LLM-based metrics.} across various NLG tasks, including dialogue, summarization, and question generation.

\begin{figure*}
    \centering
    \includegraphics[width=0.99 \linewidth]{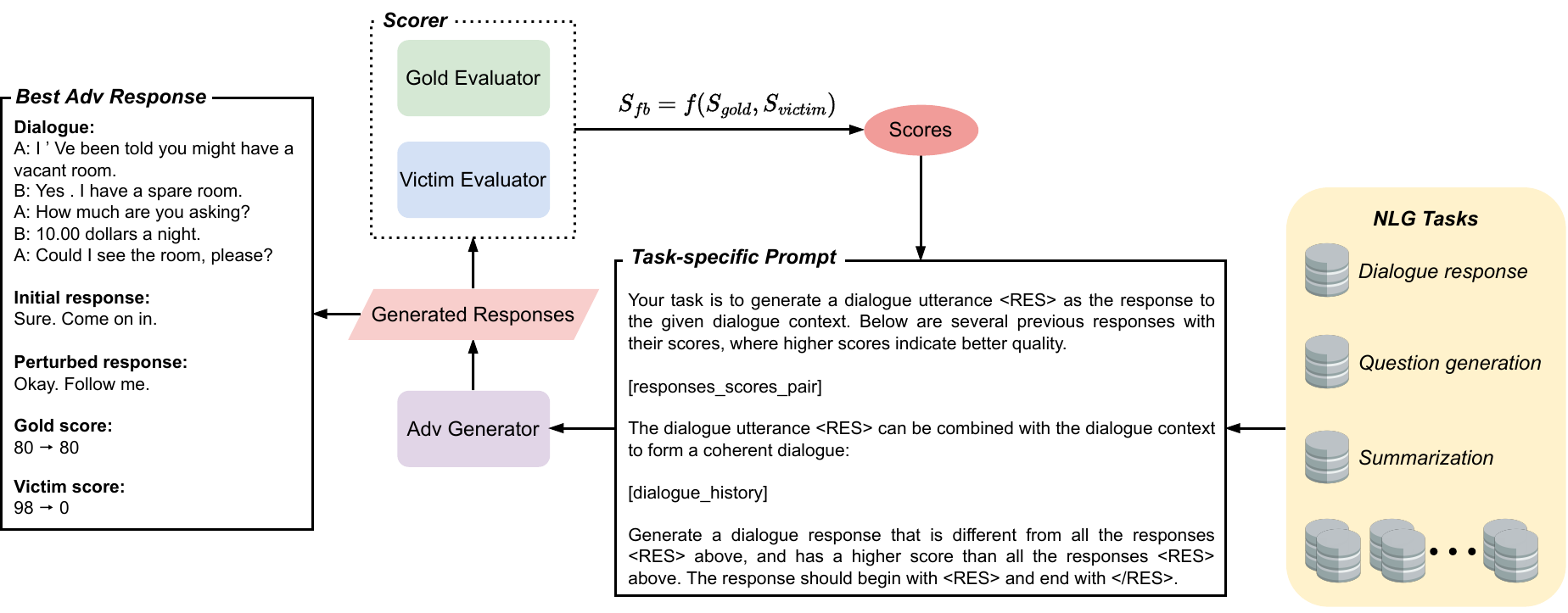}
    \caption{Overall framework of AdvEval. Optimization prompt shown above is for dialogue response generation.}
    \label{fig:overview}
\end{figure*}

\subsection{Adversarial Attack}
Adversarial attacks are common practice for robustness evaluation through creating adversarial examples to elicit incorrect predictions from neural networks~\citep{MSH16-FD,ebrahimi-etal-2018-hotflip,LiJDLW19-textbugger,wallace-etal-2019-universal, le-etal-2022-perturbations, cheng2020seq2sick,zhang-etal-2021-crafting}.
These attacks manipulate benign inputs at different granularity levels. Mainstream perturbations include character-level~\cite{belinkov2017synthetic,ebrahimi-etal-2018-adversarial}, word-level~\citep{ren-etal-2019-generating,Jin_Jin_Zhou_Szolovits_2020,li-etal-2021-contextualized,garg-ramakrishnan-2020-bae}, phrase-level~\citep{lei-etal-2022-phrase,chen-etal-2021-multi,zheng-etal-2020-evaluating}, and sentence-level~\citep{wang-etal-2020-t3,lin-etal-2021-using,xu-etal-2021-grey}.
The fundamental assumption of these works is label preservation under limited minor perturbation. The assumption does hold for various classification tasks, e.g., sentiment and intent classification. Yet, these perturbations are highly likely to affect the text quality, which is critical in NLG evaluation tasks. 
This discrepancy renders traditional adversarial techniques unsuitable for attacking NLG evaluators.

While there are also adversarial works targeting NLG evaluators~\citep{khalid-lee-2022-explaining, sai-etal-2021-perturbation}, they bypass the label concern by using only heuristic perturbation rules. These rules include invariant rules, e.g., phrase expansion/contraction, and deterioration rules, e.g., dropping words and reordering sentences. Yet, these rules are relatively simple and follow a certain pattern, rendering them infeasible to scale up and create diverse, high-quality adversarial data. In contrast, AdvEval utilizes a powerful generative model to craft adversarial samples with automatic quality control, which can generate large-scale, diverse natural adversarial samples.

\section{Methodology}

\subsection{Problem Formulation}
Our study focuses on conditional generation tasks, where a single data point consists of input context $C$, output response $R$, and reference ground-truth response $G$. Given the context and response, the evaluator $E$ will assign a score $S$, indicating the quality of the response w.r.t. context:
\begin{equation}
    S = E(C, R, G).
\end{equation}
We aim to design a generalizable framework to generate adversarial samples for various NLG evaluation tasks. Given a victim evaluator $E_{victim}$, we target to obtain a perturbed response from initial response $R_{0}$ that leads to the disagreement between score $S_{victim}$ from $E_{victim}$ and human judgment $S_{human}$. Specifically, the perturbed responses include two types: 1) $R^{+}$ where $S_{human}$ is high and $S_{victim}$ is low; 2) $R^{-}$ where $S_{human}$ is low and $S_{victim}$ is high. For all $E_{victim}$, we assume black-box access to all $E_{victim}$, which means only the evaluation scores are accessible.
To tackle this problem, we employ advanced LLMs as both adversarial generator $G$ and evaluator $E_{gold}$. For these two models, we only have black-box access, where only output text is available.

\subsection{Overview}
Fig.~\ref{fig:overview} illustrates the overall framework of AdvEval. The generator $G$ iteratively optimizes the adversarial response $R$. In each step, $G$ generates new candidate responses with the task-specific prompt as the input. These new candidates are then scored by both $E_{gold}$ and $E_{victim}$, generating a feedback score $S_{fb}$. Afterwards, these candidate responses, along with their feedback scores, will be included in the generation prompt for the next iteration to guide $G$ towards generating better adversarial responses. The optimization process ends if the score reaches a pre-defined threshold or the iteration count hits the maximum limit. Finally, the response with the highest score will be returned as the adversarial response. The following sections will elaborate on the task-specific prompt and scorer designs.

\begin{table*}[ht]
\centering

\resizebox{0.7\linewidth}{!}{

\begin{tabular}{lcccc}
\toprule
\textbf{Evaluator} & \textbf{Reference-free} & \textbf{Model-based} & \textbf{LLM-based} & \textbf{Studied task} \\
\midrule
SacreBLEU~\citep{post-2018-call} & & & & Dialog, Summ, QA \\
ROUGE~\citep{lin-2004-rouge} & & & & Summ \\
BLEURT~\citep{sellam-etal-2020-bleurt} & & \checkmark & & Dialog, QA \\
BERTScore~\citep{Zhang*2020BERTScore:} & & \checkmark & & Dialog, Summ \\
PoE~\citep{zhang2023poe} & \checkmark & \checkmark & & Dialog \\
UniEval~\citep{zhong-etal-2022-towards} & \checkmark & \checkmark & & Dialog, Summ \\
BARTScore~\citep{yuan2021bartscore} & \checkmark & \checkmark & & Summ \\
TRUE~\citep{honovich-etal-2022-true} & \checkmark & \checkmark & & Summ \\
RQUGE~\citep{mohammadshahi-etal-2023-rquge} & \checkmark & \checkmark & & QA \\
Baichuan~\citep{baichuan2023baichuan2} & \checkmark & \checkmark & \checkmark & Dialog, QA \\
Vicuna~\citep{zheng2023judging} & \checkmark & \checkmark & \checkmark & Dialog, QA \\
InstructGPT~\citep{ouyang2022training} & \checkmark & \checkmark & \checkmark & Dialog \\
\bottomrule
\end{tabular}

}
\caption{Summary of victim models studied in AdvEval.}
\label{tab:victim-models}
\end{table*}

\subsection{Generation Prompt}
We design task-specific prompts to iteratively improve the generated adversarial samples. Fig.~\ref{fig:overview} shows a dialogue task example prompt. The task-specific prompt includes below components.

\textbf{Task instruction:} We give instructions to the adversarial generator about the optimization goal ("generate dialogue response with higher score"), the output format for easier parsing ("begin with <RES> and end with <RES>"), and also NLG task definition ("dialogue utterance as the response to the given dialogue context").

\textbf{Evaluation criteria:} In addition, we find that the viable response space is considerably limited, particularly for more constrained tasks like summarization. During the optimization process, the generator is prone to generate bad response, which reduces the $R^{+}$ probability. Therefore, when generating $R^{+}$, we optionally append the evaluation criteria (e.g., "The summary must be factually consistent with the article." for factual consistent summarization) to the task instruction. This emphasis makes the generator better understand the desired output, and implicitly constrain the output space.

\textbf{Optimization trajectory:} Rather than direct instructions on how to update the responses, we add optimization trajectory in the generation prompt. The optimization trajectory includes a history of generated responses and their associated scores, organized in ascending score order. The score serves as the proxy of disagreement between human and victim evaluator (Sec.~\ref{sec:human-proxy}). We only keep top-10 responses with highest scores to reduce the context length and inference cost. We aims to use the optimization trajectory to guide $G$ towards improved responses by combining existing good responses, or perturbing the responses along a potential desired direction

\subsection{Human Judgement Proxy}
\label{sec:human-proxy}
As mentioned previously, our optimization goal here is to maximize disagreement between $S_{victim}$ and $S_{human}$. The disagreement serves as both the criterion for successful attacks and crucial feedback to the adversarial generator. Yet, unlike previous work on prompt optimization, where the goal function is simply accuracy and easily accessible, it's costly and infeasible to ask human to participate in the optimization process, complicating the assignment of scores that accurately reflect human judgment.
To address this problem, we propose to use a strong evaluator $E_{gold}$ to produce $S_{gold}$ as a proxy of $S_{human}$. Specifically, the $E_{gold}$ can be a single LLM-based evaluator or an ensemble of several powerful evaluators. The $E_{gold}$ is asked to rate the response on a 100-point scale using the manual/automatic prompt shown in Appx.~\ref{sec:appendix-eval-prompt}. Since we observe that LLM-based evaluators sometimes produce unstable scores, we utilize self-agreement to produce a consistent score. We first obtain 8 ratings from the evaluator through repeatedly sampling. Then, we take the arithmetic mean of these ratings as the final gold score $S_{gold}$. After getting the $S_{gold}$, it is combined with $S_{victim}$ to get the final score $S_{fb}$ as the feedback to generator:
\begin{equation}
S_{fb} = \begin{cases}
\alpha \times S_{gold}-S_{victim},\quad&\text{if}\ R^{+} \\
S_{victim}-\alpha \times S_{gold},\quad&\text{if}\ R^{-}
\end{cases},
\end{equation}
where $\alpha$ is hyper-parameter balancing the influence of $S_{gold}$ and $S_{victim}$. A higher $\alpha$ encourages the generator to exploit the response space adhere to judgement of gold evaluator, while a lower $\alpha$ allows the generator to more aggressively explore diverse responses that can lead to misjudgement of victim evaluator.

\begin{table*}[ht]
\centering

\resizebox{\textwidth}{!}{

\begin{tabular}{lcccccccccccccc}
\toprule
\textbf{Model} & \multicolumn{7}{c}{\textbf{DailyDialog}} & \multicolumn{7}{c}{\textbf{DREAM}} \\
\cmidrule(lr){2-8} \cmidrule(lr){9-15} 
 & SR & EX & JB & NU & BA & TF & Ours & SY & EX & JB & NU & BA & TF & Ours \\
\midrule
SacreBLEU & 3 / -- & 4 / -- & -- / -- & -- / -- & 20 / -- & 29 / -- & \textbf{100 / --} & 5 / -- & 2 / -- & -- / -- & -- / -- & 10 / -- & 17 / -- & \textbf{100 / --} \\
BLEURT & 0 / -- & 0 / -- & -- / -- & -- / -- & 1 / -- & 6 / -- & \textbf{79 / --} & 0 / -- & 0 / -- & -- / -- & -- / -- & 0 / -- & 3 / -- & \textbf{92 / --} \\
BERTScore & 0 / -- & 0 / -- & -- / -- & -- / -- & 4 / -- & 16 / -- & \textbf{93 / --} & 0 / -- & 0 / -- & -- / -- & -- / -- & 6 / -- & 8 / -- & \textbf{99 / --} \\
PoE-Base & 10 / -- & 13 / -- & -- / 39 & -- / 66 & 53 / 35 & 63 / 13 & \textbf{99 / 95} & 18 / -- & 8 / -- & -- / 2 & -- / 78 & 39 / 58 & 40 / 35 & \textbf{90 / 96} \\
PoE-Large & 6 / -- & 10 / -- & -- / 49 & -- / 72 & 54 / 46 & 52 / 25 & \textbf{95 / 98} & 8 / -- & 4 / -- & -- / 2 & -- / 77 & 28 / 63 & 39 / 34 & \textbf{81 / 97} \\
UniEval & 0 / -- & 3 / -- & -- / 22 & -- / 76 & 55 / 69 & 27 / 74 & \textbf{95 / 96} & 5 / -- & 4 / -- & -- / 4 & -- / 75 & 31 / 74 & 34 / 41 & \textbf{98 / 96} \\
Baichuan-7B & 6 / -- & 12 / -- & -- / 34 & -- / 24 & 56 / 68 & 43 / 51 & \textbf{95 / 94} & 16 / -- & 19 / -- & -- /34 & -- / 30 & 32 / 69 & 31 / 59 & \textbf{90 / 88} \\
Baichuan-13B & 0 / -- & 3 / -- & -- / 27 & -- / 41 & 57 / 76 & 16 / \textbf{98} & \textbf{76} / 97 & 2 / -- & 8 / -- & -- / 27 & -- / 45 & 19 / 83 & 21 / \textbf{100} & \textbf{73} / 98 \\
Vicuna-7B & 6 / -- & 9 / -- & -- / 25 & -- / 14 & 58 / 63 & 43 / 36 & \textbf{98 / 88} & 9 / -- & 14 / -- & -- / 22 & -- / 27 & 36 / 59 & 33 / 37 & \textbf{92 / 84} \\
Vicuna-13B & 7 / -- & 6 / -- & -- / 20 & -- / 12 & 59 / 42 & 44 / 31 & \textbf{92 / 88} & 8 / -- & 16 / -- & -- / 17 & -- / 10 & 34 / 50 & 40 / 32 & \textbf{90 / 87} \\
\midrule
Avg. & 3.8 / -- & 6 / -- & -- / 30.9 & -- / 43.6 & 41.7 / 57.0 & 33.9 / 46.9 & \textbf{92.2 / 93.7} & 7.1 / -- & 7.5 / -- & -- / 12.3 & -- / 48.9 & 23.5 / 65.1 & 26.6 / 48.3 & \textbf{90.5 / 92.3} \\
\bottomrule
\toprule
 & \multicolumn{7}{c}{\textbf{MuTual}} & \multicolumn{7}{c}{\textbf{PersonaChat}} \\
 \cmidrule(lr){2-8} \cmidrule(lr){9-15} 
 & SY & EX & JB & NU & BA & TF & Ours & SY & EX & JB & NU & BA & TF & Ours \\
 \midrule
SacreBLEU & 3 / -- & 4 / -- & -- / -- & -- / -- & 8 / -- & 10 / -- & \textbf{100 / --} & 2 / -- & 2 / -- & -- / -- & -- / -- & 2 / -- & 2 / -- & \textbf{100 / --} \\
BLEURT & 0 / -- & 0 / -- & -- / -- & -- / -- & 0 / -- & 4 / -- & \textbf{81 / --} & 0 / -- & 0 / -- & -- / -- & -- / -- & 1 / -- & 1 / -- & \textbf{69 / --} \\
BERTScore & 0 / -- & 0 / -- & -- / -- & -- / -- & 4 / -- & 3 / -- & \textbf{85 / --} & 0 / -- & 0 / -- & -- / -- & -- / -- & 1 / -- & 1 / -- & \textbf{85 / --} \\
PoE-Base & 12 / -- & 7 / -- & -- / 0 & -- / 84 & 33 / 51 & 39 / 23 & \textbf{92 / 96} & 20 / -- & 9 / -- & -- / 0 & -- / 80 & 3 / 33 & 7 / 32 & \textbf{77 / 100} \\
PoE-Large & 5 / -- & 4 / -- & -- / 1 & -- / 84 & 16 / 60 & 39 / 26 & \textbf{74 / 99} & 17 / -- & 8 / -- & -- / 1 & -- / 87 & 3 / 36 & 8 / 44 & \textbf{72 / 97} \\
UniEval & 2 / -- & 5 / -- & -- / 5 & -- / 74 & 17 / 74 & 13 / 39 & \textbf{96 / 97} & 13 / -- & 9 / -- & -- / 9 & -- / 84 & 3 / 40 & 8 / 45 & \textbf{92 / 94} \\
Baichuan-7B & 7 / -- & 10 / -- & -- / 26 & -- / 41 & 24 / 73 & 25 / 75 & \textbf{89 / 91} & 17 / -- & 12 / -- & -- / 14 & -- / 23 & 9 / 41 & 11 / 26 & \textbf{80 / 93} \\
Baichuan-13B & 2 / -- & 4 / -- & -- / 32 & -- / 48 & 13 / 84 & 20 / \textbf{100} & \textbf{73} / 99 & 11 / -- & 6 / -- & -- / 25 & -- / 49 & 3 / 46 & 0 / \textbf{100} & \textbf{52} / 97 \\
Vicuna-7B & 4 / -- & 9 / -- & -- / 11 & -- / 26 & 22 / 62 & 26 / 23 & \textbf{92 / 88} & 23 / -- & 13 / -- & -- / 4 & -- / 5 & 4 / 26 & 4 / 9 & \textbf{81 / 73} \\
Vicuna-13B & 6 / -- & 14 / -- & -- / 8 & -- / 12 & 19 / 42 & 28 / 18 & \textbf{88 / 88} & 25 / -- & 14 / -- & -- / 5 & -- / 2 & 5 / 12 & 10 / 8 & \textbf{85 / 75} \\
\midrule
Avg. & 4.1 / -- & 5.7 / -- & -- / 11.9 & -- / 51.7 & 15.6 / 63.7 & 20.7 / 43.4 & \textbf{87 / 94} & 12.8 / -- & 7.3 / -- & -- / 8.3 & -- / 47.1 & 3.4 / 33.4 & 5.2 / 37.7 & \textbf{79.3 / 89.9} \\
\bottomrule
\end{tabular}

}
\caption{Experiment results on dialogue response evaluation. Each row refers to different victim evaluators, and each column refers to different baseline methods: synonym replacement (SR), extension (EX), jumble (JB), negate previous utterance (NU), BERT-Attack (BA), TextFooler (TF). Each cell gives the $R^{+}$ ASR (left) and $R^{-}$ ASR (right) respectively. For those methods that are designed to generate only single type of adversarial data, we report only one ASR score. We also omit $R^{-}$ ASR for reference-based metrics, since adding any perturbation will lead to deteriorated score. Results with highest ASR are bold.}
\label{tab:main-dialog}
\end{table*}

\section{Experiments}
\subsection{Experiment Setup}
Due to space limitations, for more details about model implementation and hyper-parameters, please refer to Appx.~\ref{sec:appendix-experiment-setup}.
\label{sec:experiment-setup}

\textbf{Datasets:} We conduct our experiments on three popular NLG evaluation tasks: dialogue response, summarization, and question evaluation. We assess proposed AdvEval with 100 samples from each dataset. For dialogue response, we utilize DailyDialog~\citep{li-etal-2017-dailydialog}, DREAM~\citep{sun-etal-2019-dream}, PersonaChat~\citep{zhang-etal-2018-personalizing}, and MuTual~\citep{cui-etal-2020-mutual}. For summarization, data come from CNN/DailyMail~\citep{see-etal-2017-get}, XSum~\cite{narayan-etal-2018-dont}, and DialogSum~\citep{chen-etal-2021-dialogsum}. For question generation, we use SQuAD~\citep{rajpurkar-etal-2016-squad}, NewsQA~\citep{trischler-etal-2017-newsqa}, HotpotQA~\citep{yang-etal-2018-hotpotqa} and Natural Questions~\citep{kwiatkowski-etal-2019-natural}.

\textbf{Evaluation dimensions:} In our study, each NLG task focuses on a key evaluation dimension. For dialogue response, we emphasize relevance, which strongly correlates with human judgment. For the summarization task, we prioritize factual consistency, a vital aspect of this task. Lastly, for question evaluation, we assess overall quality, encompassing both the answer-ability of the questions and their relevance to the provided answers.

\textbf{Adversarial generators \& gold evaluators:} 
For consistency and simplicity, we utilize the same LLM as gold evaluator and adversarial generator. For dialogue response generation, we use PALM-2. For summarization and question generation, PALM-2 experiences severe hallucination, and fails to generate factual consistent summaries or answerable questions in our preliminary experiments. Therefore, we use GPT-4 as the $R^{+}$ generator/evaluator, and PALM-2 as the $R^{-}$ generator/evaluator.

\textbf{Victim evaluators:} Our study involves a thorough examination of a diverse array of evaluators, which vary in mechanism and model size. Tab.~\ref{tab:victim-models} lists these evaluators.\footnote{Experiments on InstructGPT is presented in APPX.~\ref{sec:appendix-attack-instructgpt}.}

\begin{table*}[ht]
\centering

\resizebox{\textwidth}{!}{
\begin{tabular}{lcccccccccccccc}
\toprule
\textbf{Model} & \multicolumn{7}{c}{\textbf{HotpotQA}} & \multicolumn{7}{c}{\textbf{NaturalQuestion}} \\
 \cmidrule(lr){2-8} \cmidrule(lr){9-15}
\textbf{} & SR & CO & CA & CW & BA & TF & Ours & SR & CO & CA & CW & BA & TF & Ours \\
\midrule
SacreBLEU & 2 / -- & 0 / -- & -- / -- & -- / -- & 21 / -- & 35 / -- & \textbf{100 / --} & 2 / -- & 0 / -- & -- / -- & -- / -- & 29 / -- & 35 / -- & \textbf{100 / --} \\
BLEURT & 0 / -- & 0 / -- & -- / -- & -- / -- & 1 / -- & 9 / -- & \textbf{95 / --} & 1 / -- & 0 / -- & -- / -- & -- / -- & 7 / -- & 10 / -- & \textbf{99 / --} \\
RQUGE & 5 / -- & 4 / -- & -- / 33 & -- / 59 & 63 / 28 & 64 / 5 & \textbf{77 / 89} & 6 / -- & 8 / -- & -- / 34 & -- / 44 & 47 / 47 & 51 / 9 & \textbf{96 / 73} \\
Baichuan-7B & 0 / -- & 0 / -- & -- / 65 & -- / 84 & 11 / 92 & \textbf{13} / 92 & 11 / \textbf{100} & 1 / -- & 2 / -- & -- / 50 & -- / 66 & 9 / 94 & \textbf{14} / 95 & \textbf{14 / 100} \\
Baichuan-13B & 3 / -- & 0 / -- & -- / 56 & -- / 67 & 25 / 72 & 34 / 13 & \textbf{52 / 99} & 0 / -- & 1 / -- & -- / 45 & -- / 52 & 27 / 78 & \textbf{35} / 21 & 25 / \textbf{100} \\
Vicuna-7B & 6 / -- & 3 / -- & -- / 13 & -- / 37 & 28 / 35 & 34 / \textbf{100} & \textbf{61} / 96 & 7 / -- & 10 / -- & -- / 5 & -- / 17 & 26 / 31 & 41 / \textbf{100} & \textbf{53} / 98 \\
Vicuna-13B & 5 / -- & 2 / -- & -- / 47 & -- / 54 & 39 / 47 & 40 / 5 & \textbf{60 / 98} & 3 / -- & 5 / -- & -- / 41 & -- / 59 & 25 / 33 & \textbf{44} / 1 & 35 / \textbf{98} \\
\midrule
Avg. & 3 / -- & 1.29 / -- & -- / 42.8 & -- / 60.2 & 26.9 / 54.8 & 32.7 / 43 & \textbf{65.1 / 96.4} & 2.86 / -- & 3.71 / -- & -- / 35 & -- / 47.6 & 24.3 / 56.6 & 32.9 / 45.2 & \textbf{60.3 / 93.8} \\
\bottomrule
\toprule
 & \multicolumn{7}{c}{\textbf{NewsQA}} & \multicolumn{7}{c}{\textbf{SQUAD}} \\
 \cmidrule(lr){2-8} \cmidrule(lr){9-15}
 & SR & CO & CA & CW & BA & TF & Ours & SR & CO & CA & CW & BA & TF & Ours \\
 \midrule
SacreBLEU & 2 / -- & 4 / -- & -- / -- & -- / -- & 44 / -- & 55 / -- & \textbf{100 / --} & 1 / -- & 2 / -- & -- / -- & -- / -- & 49 / -- & 50 / -- & \textbf{100 / --} \\
BLEURT & 0 / -- & 0 / -- & -- / -- & -- / -- & 7 / -- & 13 / -- & \textbf{95 / --} & 0 / -- & 0 / -- & -- / -- & -- / -- & 3 / -- & 16 / -- & \textbf{98 / --} \\
RQUGE & 0 / -- & 3 / -- & -- / 63 & -- / 64 & 29 / 50 & 44 / 15 & \textbf{80 / 68} & 2 / -- & 0 / -- & -- / 67 & -- / 79 & 64 / 55 & 57 / 10 & \textbf{94 / 81} \\
Baichuan-7B & 1 / -- & 1 / -- & -- / 86 & -- / 91 & 5 / 92 & \textbf{16} / 48 & 12 / \textbf{100} & 1 / -- & 1 / -- & -- / 50 & -- / 80 & 20 / 88 & \textbf{34} / 98 & 25 / \textbf{99} \\
Baichuan-13B & 0 / -- & 0 / -- & -- / 86 & -- / 61 & 15 / 85 & \textbf{27} / 88 & 14 / \textbf{98} & 0 / -- & 0 / -- & -- / 82 & -- / 78 & \textbf{44}\textbf{} & 37 / 18 & 33 / \textbf{98} \\
Vicuna-7B & 1 / -- & 7 / -- & -- / 15 & -- / 27 & 45 / 34 & 50 / \textbf{95} & \textbf{67} / 85 & 3 / -- & 2 / -- & -- / 21 & -- / 46 & 43 / 37 & \textbf{48 / 100} & 26 / 92 \\
Vicuna-13B & 1 / -- & 1 / -- & -- / 53 & -- / 69 & 29 / 41 & 41 / 12 & \textbf{42 / 87} & 0 / -- & 0 / -- & -- / 65 & -- / 79 & 41 / 44 & \textbf{46} / 5 & 32 / \textbf{94} \\
\midrule
Avg. & 0.7 / -- & 2.3 / -- & -- / 60.6 & -- / 62.4 & 24.9 / 60.4 & 35.1 / 51.6 & \textbf{58.6 / 87.6} & 1 / -- & 0.71 / -- & -- / 57 & -- / 72.4 & 37.7 / 61.2 & 41.1 / 46.2 & \textbf{58.3 / 92.8} \\
\bottomrule
\end{tabular}

}
\caption{Experiment results on question evaluation. Baselines include: synonym replacement (SR), contraction (CO), change question to assertion (CA), change question word (CW), BERT-Attack (BA), TextFooler (TF).}
\label{tab:main-qg}
\end{table*}

\textbf{Metrics:} Our primary metric for evaluating adversarial methods is the attack success rate (ASR). We generate two types of adversarial data, and a successful attack must satisfy specific criteria:
\begin{equation}
\label{eq:asr-criteria}
\begin{aligned}
    &S_{gold}>\tau_{1}\ \land\ S_{gold}-S_{victim}>\tau_{2}, \text{if}\ R^{+} \\
    &S_{gold}<\tau_{1}\ \land\ S_{victim}-S_{gold}>\tau_{2}, \text{if}\ R^{-}
\end{aligned},
\end{equation}
where $\tau_{1}$ and $\tau_{2}$ are pre-defined thresholds. The first part of each condition ensures the adversarial data has a confident label (positive/negative), while the second part ensures the incorrect prediction by victim evaluators. Note that our ASR criteria involve $E_{gold}$, which is a neural metric. Therefore, on top of ASR, we also manually check the label validity of the adversarial data for trustworthy scoring. We report the rate of data with correct labels.

\textbf{Baselines:} We compare AdvEval against two types of baselines. The first type of baselines are specially designed for NLG evaluators, and utilize sophisticated rule-based perturbations to ensure the label validity~\cite{sai-etal-2021-perturbation,khalid-lee-2022-explaining}. The perturbations include invariant perturbation (synonym adjective, expansion, contraction) for $R^{+}$ generation, and deterioration perturbation (jumble, change name, negate previous utterance, change question to assertion, change question word) for $R^{-}$ generation. The first type of baselines serve as our major baseline for their high relevance. For the second type of baselines, we adapt existing attack methods originally designed for general classification tasks, including TextFooler~\citep{Jin_Jin_Zhou_Szolovits_2020} and BERT-Attack~\citep{li-etal-2020-bert-attack}, for targeted attack with target label 0 to generate $R^{+}$ and target label 1 to generate $R^{-}$. We sample a random response as the initial response during generating $R^{-}$ to align with the label preservation assumption in these methods.
For fair comparison, we allocate the same query budget 300 to both AdvEval and the baseline methods.

\subsection{Main Results}
\textbf{Dialogue response evaluation:} The comparison of different attacking methods against dialogue evaluation metrics are shown in Tab.~\ref{tab:main-dialog}. Overall, we have the following key observations. 
First, all evaluated metrics, regardless of their working mechanisms, are vulnerable to adversarial perturbations. Especially for $R^{-}$ case, even simple heuristic rule-based perturbations can deceive some models effectively. For example, negating previous utterance achieves 47.8\% ASR on average. 
Second, despite the promising performance of rule-based methods on generating $R^{-}$, their effectiveness in creating $R^{+}$ is limited, with only ASR 6.9\% with synonym replacement, and 6.6\% with phrase extension. These methods maintain label accuracy but lack data diversity and are easily detectable.
Third, compared to rule-based methods, optimization-based baselines are more effective for $R^{+}$ generation, but fall short in generating $R^{-}$. Additionally, it struggles to produce adversarial samples for reference-based methods, which often assign high scores to perturbed responses due to the high vocabulary overlap with the reference caused by local perturbations.
Lastly, AdvEval significantly outperforms all baselines, achieving an ASR of 84.8\% for $R^{+}$ and 92.5\% for $R^{-}$. This underscores the efficiency of AdvEval in navigating both $R^{+}$ and $R^{-}$ scenarios. In addition, we observe that it's easier to craft $R^{-}$ than $R^{+}$ for all attacking methods, likely due to limited good response space compared to bad response space.

\begin{table}[!ht]
\centering

\resizebox{\columnwidth}{!}{
\begin{tabular}{lcccccc}
\toprule 
\textbf{Model} & \multicolumn{6}{c}{\textbf{CNN/Dailymail}} \\
\cmidrule(lr){2-7}
\textbf{}& SR & JB& CN& BA& TF & Ours \\
\midrule
SacreBLEU& 11 / --& -- / -- & -- / -- & 0 / --& 6 / -- & \textbf{100 / --}\\
ROUGE& 0 / -- & -- / -- & -- / -- & 0 / --& 20 / --& \textbf{100 / --}\\
BERTScore& 3 / -- & -- / -- & -- / -- & 1 / --& 1 / -- & \textbf{100 / --}\\
BARTScore& 15 / --& -- / 0& -- / 18 & 69 / 3& 76 / 1 & \textbf{84 / 40} \\
UniEval& 4 / -- & -- / 11 & -- / 9& 18 / 30 & 29 / 7 & \textbf{68 / 80} \\
TRUE & 6 / -- & -- / 62 & -- / 2& 42 / 8& 53 / 2 & \textbf{72 / 41} \\
\midrule
Avg. & 6.5 / -- & -- / 24.3 & -- / 9.7& 21.7 / 13.7 & 30.8 / 3.3 & \textbf{87.3 / 53.7} \\
\toprule
\bottomrule
 & \multicolumn{6}{c}{\textbf{XSum}}\\
 \cmidrule(lr){2-7}
 & SR & JB& CN& BA& TF & Ours \\
 \midrule
SacreBLEU& 7 / -- & -- / -- & -- / -- & 6 / --& 20 / --& \textbf{100 / --}\\
ROUGE& 0 / -- & -- / -- & -- / -- & 1 / --& 4 / -- & \textbf{100 / --}\\
BERTScore& 0 / -- & -- / -- & -- / -- & 1 / --& 4 / -- & \textbf{100 / --}\\
BARTScore& 23 / --& -- / 0& -- / 20 & 58 / 25 & 68 / 3 & \textbf{81 / 35} \\
UniEval& 10 / --& -- / 0& -- / 12 & 60 / 45 & 23 / 6 & \textbf{63 / 60} \\
TRUE & 17 / --& -- / 11 & -- / 1& 71 / 1& 57 / 1 & \textbf{85 / 43} \\
\midrule
Avg. & 9.5 / -- & -- / 3.7& -- / 11 & 32.8 / 23.7 & 29.3 / 3.3 & \textbf{88.2 / 46} \\
\bottomrule
\toprule
 & \multicolumn{6}{c}{\textbf{DialogSum}} \\
\cmidrule(lr){2-7}
 & SR & JB& CN& BA& TF & Ours \\
\midrule
SacreBLEU& 6 / -- & -- / -- & -- / -- & 11 / -- & 17 / --& \textbf{100 / --}\\
ROUGE& 2 / -- & -- / -- & -- / -- & 5 / --& 3 / -- & \textbf{100 / --}\\
BERTScore& 0 / -- & -- / -- & -- / -- & 3 / --& 6 / -- & \textbf{97 / --} \\
BARTScore& 22 / --& -- / 0& -- / 40 & 64 / 23 & 69 / 1 & \textbf{80 / 43} \\
UniEval& 0 / -- & -- / 21 & -- / 21 & 45 / 79 & \textbf{48} / 13 & 33 / \textbf{99} \\
TRUE & 5 / -- & -- / 53 & -- / 15 & 49 / 1& 65 / 0 & \textbf{67 / 88} \\
\midrule
Avg. & 5.8 / -- & -- / 24.7 & -- / 25.3 & 29.5 / 34.3 & 34.7 / 4.7 & \textbf{79.5 / 76.7} \\
\bottomrule
\end{tabular}

}
\caption{Experiment results on summarization evaluation. Baselines include: synonym replacement (SR), jumble (JB), change name (CN), BERT-Attack (BA), TextFooler (TF).}
\label{tab:main-sum}
\end{table}


\textbf{Question evaluation:} 
The results on question evaluation are shown in Tab.~\ref{tab:main-qg}. Findings from the question evaluation largely mirror those from the dialogue evaluation tasks. AdvEval achieves average $R^{+}$ success rate 60.6\% and $R^{-}$ success rate 92.7\%. However, we observe a lower $R^{+}$ ASR compared to dialogue tasks. The lower $R^{+}$ ASR might be attributed to the stringent constraint, i.e., questions to be answerable from the article and aligned with the provided answer. This leads to a limited range of appropriate responses. Despite this, AdvEval still outperforms baseline methods in average performance.

\textbf{Summarization evaluation:} The results on summarization evaluation are shown in Tab.~\ref{tab:main-sum}. AdvEval surpasses baseline methods with an 85\% $R^{+}$ ASR, and 58.8\% $R^{-}$ ASR. In addition, these summarization metrics are more robust against $R^{-}$ perturbation compared to dialogue and question metrics. The highest rule-based average ASR is only 15.3\%, in stark contrast to 60.2\% for question metric, and 47.8\% for dialogue metric.

Overall, our proposed AdvEval significantly outperforms existing adversarial techniques across all examined tasks. The superior performance of AdvEval can be attributed to two key advantages.
Firstly, AdvEval benefits from a larger search space. While all baseline methods are restricted to adding only local perturbations to the benign input, which significantly limits their search space, AdvEval leverages the remarkable generative capabilities of LLMs to perform both local and global perturbations. This results in a substantially larger search space, facilitating the discovery of novel and effective responses.
Secondly, AdvEval includes improved search constraints. Baseline methods typically rely on either rule-based or sentence embedding constraints to maintain label integrity. However, as discussed in our paper, these constraints fall short in text evaluation scenarios due to their inability to capture the nuances of natural language. To overcome this limitation, AdvEval adopts an LLM evaluator as a human proxy, which ensures superior label integrity by more accurately assessing the quality of generated responses.

\subsection{Label Validity}
In AdvEval, gold evaluator $E_{gold}$ is crucial as it acts as a human proxy for assessing response quality and guiding the adversarial generator. 
To evaluate its reliability, we conduct a manual examination on 780 adversarial data points, spanning various evaluation tasks and generated through different attack methods, with 260 data points per method.
Three annotators manually label each data point as 'good' or 'not good' based on context and scoring criteria, with majority voting determining the final label. Then, we convert the $S_{gold}$ to binary high-quality/low-quality label with threshold 50. Finally, We compute the accuracy score between $S_{gold}$ binary labels and human annotated ones to measure the reliability of $E_{gold}$ on adversarial data.
Due to space limitations, detailed annotation schema can be found in APPX.~\ref{sec:appendix-human-validation}.

As shown in Tab.~\ref{tab:manual}, $E_{gold}$ achieves an exceptionally high average accuracy over 80\%, aligning closely with the highest inter-annotator agreement rates of 77\%. This accuracy indicates a high degree of reliability of our gold evaluator, affirming its utility as a viable proxy for human judgment in guiding the optimization process. Additionally, we claim that a key issue with earlier adversarial methods is their inability to consistently preserve the quality label of $R^{+}$. This is supported by our manual labeling results that 65\% of $R^{+}$ samples from TextFooler are labelled as bad response by human annotators.

\begin{table}[ht]
\centering

\resizebox{0.62\columnwidth}{!}{

\begin{tabular}{lccc}
\toprule
\textbf{Method} & \textbf{Dialog} & \textbf{Summ} & \textbf{QA} \\
\midrule
TextFooler & 85.0 & 78.3 & 99.0 \\
BERT-Attack & 78.0 & 83.3 & 90.0 \\
\midrule
AdvEval & 75.0 & 78.3 & 72.0 \\
\bottomrule
\end{tabular}

}
\caption{Agreement between human and gold evaluator on adversarial data.}
\label{tab:manual}
\end{table}

\subsection{Adversarial Learning}
To assess potential countermeasures against adversarial attacks, we experiment with the adversarial learning on PoE-Large. We generate 3,649 adversarial data points from mixed data source of DREAM, DailyDialog, and MuTual with AdvEval, and also 3,634 adversarial data points with TextFooler. Then, we further finetune the model on these two types of adversarial data mixed with original vanilla data respectively. As shown in Tab.~\ref{tab:adv-poe}, TextFooler can be easily defended with naive adversarial training, while our proposed AdvEval is more robust against this defense technique. After adversarial training, the model can hardly achieve any improvement, which further validates the capacity of AdvEval. We hypothesize that AdvEval can craft more diverse and complex data points that challenge the model's adaptability.

\begin{table}[ht]
\centering

\resizebox{0.88\columnwidth}{!}{

\begin{tabular}{lccc}
\toprule
\textbf{Method} & \textbf{DailyDialog} & \textbf{DREAM} & \textbf{MuTual} \\
\midrule
AdvEval & 95 / 98& 81 / 97& 74 / 99 \\
+Adv. Train & 93 / 98& 81 / 97& 74 / 99 \\
\midrule
TextFooler& 52 / 27& 39 / 44& 39 / 32 \\
+Adv. Train & 9 / 6& 6 / 3& 5 / 8 \\
\bottomrule
\end{tabular}

}
\caption{Adversarial learning on PoE-Large using data crafted by different methods.}
\label{tab:adv-poe}
\end{table}

\begin{table}[!ht]
\centering

\small
\begin{tabularx}{0.98 \columnwidth}{X}
\toprule
\multicolumn{1}{c}{\textit{Dialog Response Evaluation}} \\
\midrule
A: I'd like to taste some local dishes . What would you recommend ? \\
B: That's fine . You must try this dish . \\
A: Could you tell me how this thing is cooked ? \\
B: It's fish steamed and served with our special sauce . \\
A: Is it good ? \\
\midrule
\textbf{{[}Benign $\bm{R}${]}} Sure, It’s a most popular dish. \\
\midrule
\textbf{{[}AdvEval $\bm{R^{+}}${]}} You'll never eat better. \\
\textbf{{[}AdvEval $\bm{R^{-}}${]}} I have never had it. \\
\midrule
\textbf{{[}BERT-Attack $\bm{R^{+}}${]}} Sure. It's a most popular day. \\
\textbf{{[}BERT-Attack $\bm{R^{-}}${]}} So, what's the bad?  \\
\midrule
\textbf{{[}TextFooler $\bm{R^{+}}${]}} Sure. It's a most popular flat. \\
\textbf{{[}TextFooler $\bm{R^{-}}${]}} So, what's the question?  \\
\bottomrule
\toprule
\multicolumn{1}{c}{\textit{Question Evaluation}} \\
\midrule
\textbf{Article:} \\
...... \\
In official Chinese histories, the Yuan dynasty bore the Mandate of Heaven, following the Song dynasty and preceding the Ming dynasty. \\
...... \\
\textbf{Answer:} \\
Ming dynasty \\
\midrule
\textbf{{[}Benign $\bm{R}${]}} What dynasty came after the Song? \\
\midrule
\textbf{{[}AdvEval $\bm{R^{+}}${]}} Which dynasty succeeded the Yuan dynasty according to the official Chinese historical records? \\
\textbf{{[}AdvEval $\bm{R^{-}}${]}} What is the only dynasty mentioned to have received the Mandate of Heaven in the text? \\
\midrule
\textbf{{[}BERT-Attack $\bm{R^{+}}${]}} What dynasty came after the qing? \\
\textbf{{[}BERT-Attack $\bm{R^{-}}${]}} which ranking did Harvard find in war? \\
\midrule
\textbf{{[}TextFooler $\bm{R^{+}}${]}} What dynasty came after the Dollar? \\
\textbf{{[}TextFooler $\bm{R^{-}}${]}} What structured has Amherst unearth in 1900? \\
\bottomrule
\end{tabularx}

\caption{Adversarial data from AdvEval, BERT-Attack, and TextFooler.}
\label{tab:case-study}
\end{table}

\subsection{Case Study}
Tab.~\ref{tab:case-study} presents adversarial data from different methods. More samples can be found in Appx.~\ref{sec:appendix-case-study}. 
These samples are sampled from DailyDialog and SQUAD. Compared to previous methods, AdvEval is able to generate more valuable hard adversarial data with high quality and diversity. 
Regarding $R^{+}$ generation, BERT-Attack and TextFooler exhibit similar tendencies, notably their failure to preserve the positive label after perturbation due to the inadequacy of their perturbation techniques. 
For example, in the dialogue sample, A asks whether the dish is good, but the perturbed response answers that the flat is popular, which is not related to the dialogue context at all. 
Therefore, it should be assigned a negative label. On the contrary, AdvEval generate a good response "You'll never eat better" that directly answer the question from A. 
In addition, the AdvEval adversarial data are notably more complex. For instance, one need to infer "Ming dynasty succeeded Yuan dynasty" from "Yuan dynasty preceded Ming dynasty" to check the question's answer-ability. For $R^{-}$ generation, all methods can generate samples with correct label. Compared to random responses from TextFooler and BERT-Attack, AdvEval samples are also more challenging to validate. The case study convincingly demonstrates the limitations of current attack approaches and highlights the effectiveness and necessity of our proposed AdvEval framework.

\subsection{Ablation Study}
\textbf{Gold evaluator:} 
To understand the impact of the gold evaluator in AdvEval, we conduct a study with an AdvEval variant that only uses the victim score for feedback. For $R^{-}$ generation, we use random response as the initial input to avoid direct termination with initial input. As shown in Tab.~\ref{tab:ab-gold}, excluding the gold evaluator results in a significant drop in performance. The generation process tends to end prematurely because the score threshold is quickly met without the guidance of the gold evaluator.

\begin{table}[ht]
\centering

\resizebox{0.65\columnwidth}{!}{

\begin{tabular}{lcc}
\toprule
\textbf{Model}     & \textbf{DailyDialog} & \textbf{Dream} \\
\midrule
UniEval   & 95 / 96              & 98 / 96        \\
-$E_{gold}$     & 34 / 6               & 53 / 9         \\
\midrule
Vicuna-7B & 98 / 88              & 92 / 84        \\
-$E_{gold}$     & 33 / 4               & 34 / 2       \\ 
\bottomrule
\end{tabular}

}
\caption{Ablation study on the effect of gold evaluator for dialogue response generation.}
\label{tab:ab-gold}
\end{table}

\textbf{Evaluation criteria:} We optionally include evaluation criteria into the generation prompt, aiming to generate responses following the criteria. As shown in Tab.~\ref{tab:ab-criteria}, excluding criteria leads to lower ASR. The results validate the effect of criteria in the generation prompt for guiding the LLM to generate response more adhere to the criteria.

\begin{table}[ht]
\centering

\resizebox{0.7\columnwidth}{!}{

\begin{tabular}{lcc}
\toprule
\textbf{Method} & \textbf{CNN/Dailymail} & \textbf{DialogSum} \\
\midrule
AdvEval & 68 & 33 \\
-criteria & 61 & 28 \\
\bottomrule
\end{tabular}

}
\caption{Results on UniEval with(out) criteria included in the generation prompt. $R^{+}$ ASR is reported.}
\label{tab:ab-criteria}
\end{table}

\section{Conclusion and Future Works}
In this paper, we systematically evaluate the robustness of NLG evaluators. We also propose AdvEval, a novel black-box attack framework that effectively craft high-quality diverse adversarial samples that lead to performance degradation of various NLG evaluators. Specifically, AdvEval generates adversarial samples using LLM generator with feedback from LLM evaluator. Extensive experimental demonstrate the superior effectiveness of AdvEval across variety of NLG tasks. Future works include the development of defense methods against AdvEval, and the extension of AdvEval to other NLP tasks and NLG metrics.

\section*{Limitations}
The paper proposes the first optimization-based 
framework to examine the adversarial robustness of NLG evaluators. Yet, there are several limitations.
Firstly, the framework is only able to find absolutely bad text with score around 0 or absolutely good text with score around 100. Techniques that can generate medium-quality text with score around 50 remains unexplored. We leave it to future work. Secondly, our research primarily targets the robustness of general NLG evaluators rather than specific evaluation dimensions. The dimensions we selected for study are either overall quality or dimension that is highly correlated to the overall quality across various tasks. We leave dimension-oriented attacks as an important direction for future research.
Thirdly, adversarial attack on strong proprietary LLM evaluators, e.g., InstructGPT, demands ensemble of several powerful evaluators, which requires more computational resources. We leave a more computational efficient framework to future exploration.

\section*{Ethics Statement}
We propose an adversarial attack framework against NLG evaluators on various NLG tasks in this work. We aim to study the robustness of NLG evaluators and provide insight to inspire future works on robust NLG evaluators.
Our proposed framework may be used to attack online evaluation services. However, we believe that exploring this vulnerability and robustness of evaluator is more important than the above risks. Research studying attacks on evaluator will motivate improvements to the system security to defend against the attacks.

\section*{Acknowledgement}

This work is partly supported by ASTOUND (101071191- HORIZON- EIC -2021PATHFINDERCHALLENGES-01) funded by the European Commission, BEWORD (PID2021-126061OB-C43) funded by MCIN/AEI/10.13039/501100011033 and, as  appropriate, by “ERDF A way of making Europe”,  by the “European Union”, and the Research Grants for Young Investigators from Universidad Politécnica de Madrid (GENIUS:APOYO-JOVENES-21-TAXTYC-32K61X37) funded by Comunidad de Madrid, Shenzhen Key Laboratory of Cross-Modal Cognitive Computing funded by Shenzhen Science and Technology Innovation Program (2024) 02, Shenzhen, China.

\bibliography{anthology_1,anthology_2,custom}
\bibliographystyle{acl_natbib}

\appendix

\section{Detailed Experiment Setup}
\label{sec:appendix-experiment-setup}
\subsection{Evaluator \& Generator Implementation}
We select victim evaluators that are across diverse categories, and commonly employed in latest studies~\citep{sai2022survey}. For standard metrics like SacreBLEU, ROUGE, BLEURT, and BERTScore, we use the implementation from huggingface evaluate library\footnote{\url{github.com/huggingface/evaluate}}. For PoE, UniEval, BARTScore, TURE, RQUGE, we use the official public repositories. For open-source LLM-based evaluators like Baichuan and Vicuna, we follow the methodologies in prior studies~\citep{gupta-etal-2022-instructdial,fu2023gptscore} to compute the evaluation score directly with output logits. We use latest version of these LLMs, i.e., Baichuan2 and Vicuna-V1.5. For InstructGPT, we use latest gpt-3.5-turbo-instruct-0914. Regarding the gold evaluators and adversarial generator, the PALM-2 version we use is text-bison-001. The GPT-4 version is gpt-4-1106-preview. For all evaluation score $S$, we normalize it to the same scale of 0 to 100.

\begin{figure*}[ht]
    \centering
    \includegraphics[width=0.98\textwidth]{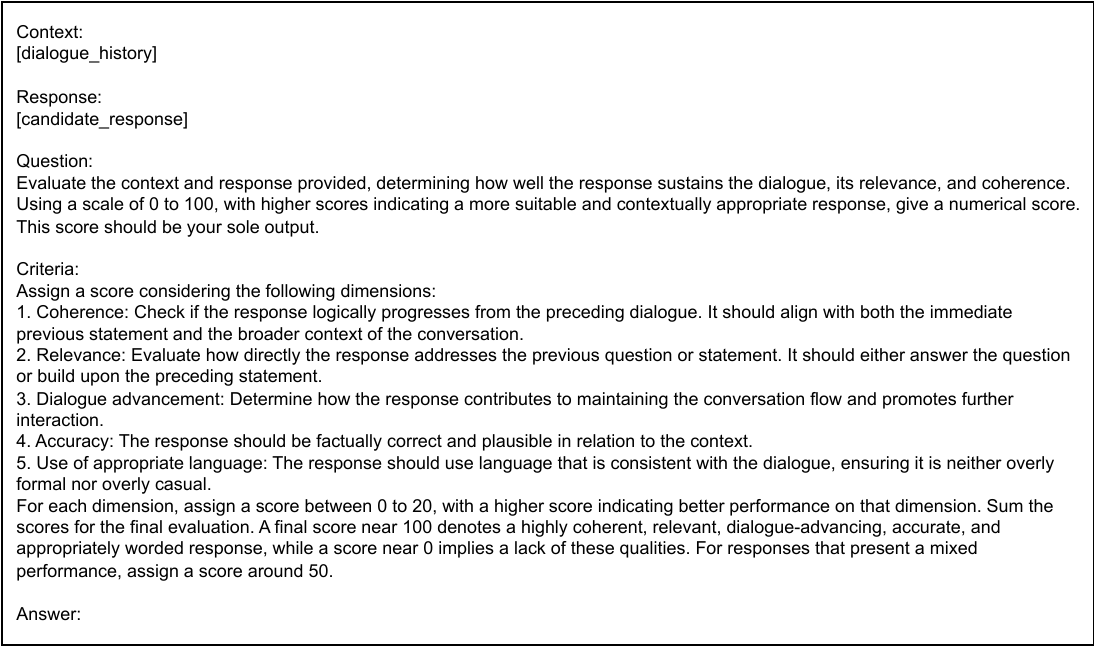}
    \caption{Example prompt for dialogue response evaluation.}
    \label{fig:eval_prompt_dialog}
\end{figure*}

\subsection{Hyper-parameters}
\label{sec:appendix-hyper-params}
Note that our objective is to get responses of distinctly high quality (with scores exceeding 70) and those of distinctly low quality (with scores below 30). Accordingly, we set $\tau_{1}$ as 70 for $R^{+}$ and 30 for $R^{-}$. $\tau_{2}$ is set at 40 across evaluations. These thresholds are chosen to ensure that adversarial samples meet two requirements: 1) distinctly representing either extremely positive or negative labels; 2) having a clear gap in scoring between the victim and gold evaluators. The second requirement implicitly ensure incorrect classification by $S_{victim}$. For simplicity, $\alpha$ is fixed at 1 across all tasks without further tuning. We observe that this simple hyper-parameter setting has demonstrated strong performance.

\begin{figure*}[!ht]
    \centering
    \includegraphics[width=0.98\textwidth]{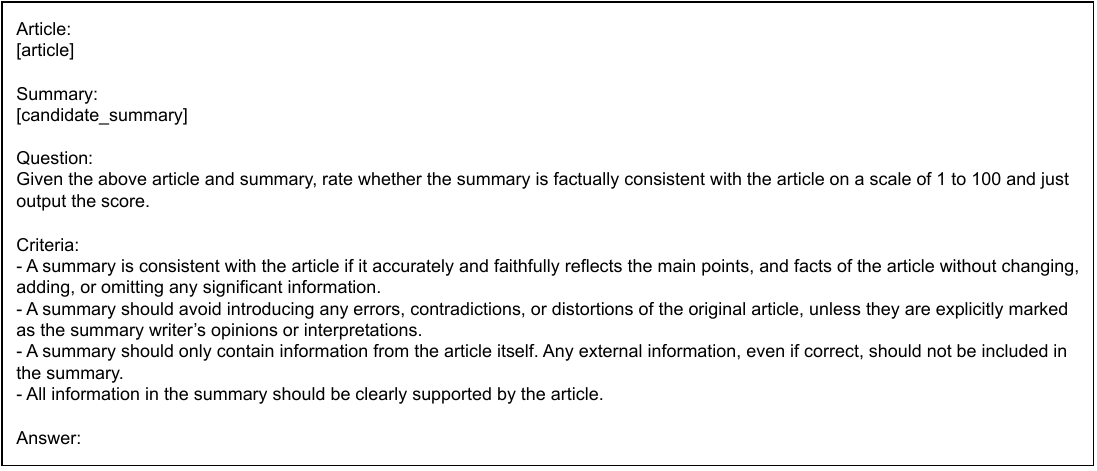}
    \caption{Example prompt for summarization evaluation.}
    \label{fig:eval_prompt_sum}
\end{figure*}

\begin{figure*}[!ht]
    \centering
    \includegraphics[width=0.98\textwidth]{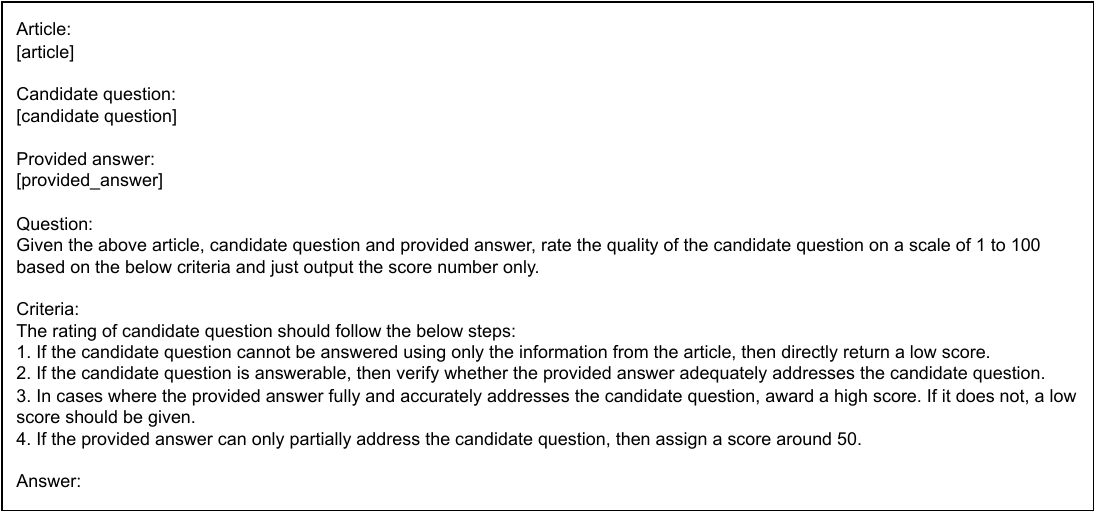}
    \caption{Example prompt for question evaluation.}
    \label{fig:eval_prompt_qa}
\end{figure*}

\section{Evaluation Prompt}
\label{sec:appendix-eval-prompt}
In this section, we list the prompt used as input to our gold evaluator (Fig.~\ref{fig:eval_prompt_dialog},~\ref{fig:eval_prompt_sum},~\ref{fig:eval_prompt_qa}). To boost the accuracy of our evaluator, the prompt can be manually crafted or automatically crafted using existing prompt optimization methods~\citep{yang2023large,liu2023calibrating} to align with human judgement. In this work, the dialogue response evaluation prompt is optimized from publicly available annotated data~\citep{zhao-etal-2020-designing,mehri-eskenazi-2020-usr,huang-etal-2020-grade}. The initial instruction provided is very straightforward and consistent with our evaluation dimension mentioned in Sec.~\ref{sec:experiment-setup}: “Given the above context and response, rate the relevance of the response to the context on a scale of 1 to 100, providing only the numerical score as your output.” The extra learned criteria can be considered as auxiliary information to support the evaluation of our targeted relevance dimension~\citep{li2024leveraging}. Similarly, We use automatically optimized summarization evaluation prompt from previous work~\citep{liu2023calibrating}. For question evaluation, we use manually crafted prompt. Note that all gold evaluation prompts explicitly specify the output format as a single score number only. We take that score number as the gold score.

\section{Attacking Proprietary LLM}
\label{sec:appendix-attack-instructgpt}
To study the possibility of using AdvEval against powerful proprietary LLMs, we further extend AdvEval to attack InstructGPT evaluator on 50 data points from DailyDialog dataset. Given the similar strong performance of available proprietary LLMs, we employ an ensemble of GPT-4 and PALM-2 as our gold evaluator. The AdvEval $R^{+}$ and $R^{-}$ ASR are 24\% and 84\% respectively. Compared to smaller victim models, InstructGPT shows superior robustness against $R^{+}$ perturbations, but remains vulnerable to $R^{-}$ perturbations. The results suggest the potential of using our proposed AdvEval framework to study the robustness issue of cutting-edge NLG evaluators, potentially guiding future improvements in evaluator robustness.

\section{Additional Case Study}
\label{sec:appendix-case-study}
\begin{table}[!ht]
\centering

\small
\begin{tabularx}{0.98 \columnwidth}{X}
\toprule
\multicolumn{1}{c}{\textit{Summarization Evaluation}} \\
\midrule
Jo: I worked hard for a whole year. I really need a break. \\
Phyllys: That's true. You need to take some time off to relax. \\
Jo: You said it. I'm looking forward to my annual vacation. \\
Phyllys: When are you going to take your vacation? \\
Jo: Later this month. I can't wait! \\
Phyllys: I really envy you. You know I'm not taking my vacation until December. \\
\midrule
\textbf{{[}Benign $\bm{R}${]}} Jo will take an annual vacation, but Phyllys cannot take it until December. \\
\midrule
\textbf{{[}AdvEval $\bm{R^{+}}${]}} Jo plans to take their well-deserved annual vacation later this month while Phyllys is scheduled to have hers in December. \\
\textbf{{[}AdvEval $\bm{R^{-}}${]}} Jo is looking forward to her annual vacation in December but Phyllys feels like her time will be wasted since she is taking her vacation then. \\
\midrule
\textbf{{[}BERT-Attack $\bm{R^{+}}${]}} Jo will take an biennial vacation, but Phyllys cannot take it until December. \\
\textbf{{[}BERT-Attack $\bm{R^{-}}${]}} Nolan didn't fail to college only. not want to skip class tomorrow to go to the movies. \\
\midrule
\textbf{{[}TextFooler $\bm{R^{+}}${]}} Qiu will take an annual vacation, but Phyllys cannot take it until December. \\
\textbf{{[}TextFooler $\bm{R^{-}}${]}} Wes didn't gonna to school fri. Brooks wished to skip echelon tommorrow to proceed to the movies. \\
\bottomrule
\end{tabularx}

\caption{Adversarial data from AdvEval, BERT-Attack, and TextFooler.}
\label{tab:case-study-appendix}
\end{table}

In Tab.~\ref{tab:case-study-appendix}, we present example adversarial data point from DialogSum with different attack techniques. Similar to the findings on dialogue and question evaluation, previous attack methods struggles in the NLG evaluation context, especially for the $R^{+}$ generation, which reinforces the necessity and impact of our approach for enhancing evaluator robustness.

\section{Human Validation of Gold Evaluator Labels}
\label{sec:appendix-human-validation}
To verify the validity of gold evaluators, we instruct PhD students who are proficient in English to annotate adversarial data produced by our AdvEval framework, as well as those from TextFooler and BERT-Attack baselines. Drawing upon manual validation scale from previous studies~\citep{mehri-eskenazi-2020-usr,yeh-etal-2021-comprehensive,mehri-eskenazi-2020-unsupervised}, a total of 780 data points are evaluated by 9 student annotators (each data point is annotated by three students). Regarding the data to be annotated, we sample adversarial data produced by AdvEval, BERT-Attack, and TextFooler. The data sources include DailyDialog, MuTual, NewsQA, SQUAD, CNN/Dailymail, and DialogSum. The victim models include BLEURT, PoE-Large, and Vicuna-13B for dialogue response; ROUGE, and UniEval for summarization; BLEURT, RQUGE, and Baichuan-7B for question generation. For each combination, we sample 10 $R^{+}$ and 10 $R^{-}$ (if reference-free).

\begin{table}[!ht]
\resizebox{0.98\linewidth}{!}{
\small
    \centering
    \colorbox{gray!8}{
    \begin{tabular}{@{}p{7.3cm}}
    ================ \textsc{Dialogue} ================\\\\
     Evaluate the context of the dialogue and the response provided, and judge the overall quality of the response in maintaining the dialogue, relevance, and coherence. It's a binary classification task, please annotate in "Yes or No." \\\\

Consider the following criteria comprehensively:\\\\
Coherence: Check if the response smoothly develops from the logic of the previous conversation. It should be consistent with the previous statement and the broader context of the dialogue. \\\\
Relevance: Evaluate how the response directly addresses the previous question or statement. It should answer the question or build upon the previous statement. \\\\
Dialogue Advancement: Determine how the response helps maintain the flow of conversation and encourages further interaction. \\\\
Accuracy: The response should be factually correct and credible in the context. \\\\
Use of Appropriate Language: The response should use language consistent with the conversation, ensuring it is neither too formal nor too casual. \\\\
============= \textsc{Summarization} =============\\\\
Based on the provided text and summary, evaluate whether the summary is factually consistent with the text. It's a binary classification task, please annotate in "Yes or No". \\\\

Consider the following criteria comprehensively:\\\\

If the summary accurately and faithfully reflects the main points and facts of the article, without changing, adding, or omitting any important information, then the summary is consistent with the article. The summary should avoid introducing any errors, contradictions, or distortions of the original text unless these are clearly marked as the views or interpretations of the author of the summary. \\\\

================== \textsc{QG} ==================\\\\

Based on the given criteria, the article, and the provided answer, evaluate the question. It's a binary classification task, please annotate in "Yes or No". \\\\

Consider the following criteria comprehensively:\\\\

If the candidate question cannot be answered using only the information in the article, then judge it as "No." \\\\
If the candidate question has an answer in the article, verify whether the provided answer addresses the candidate question. \\\\
If the provided answer accurately addresses the candidate question, judge it as "Yes." \\\\
Otherwise, judge it as "No."

\end{tabular}}
}
\caption{Instruction manual provided to the students}.
\label{tab:scoring-criteria}
\end{table}

Instead of using 100-scale annotations, We design the annotation task as a binary classification that determines whether the output text is "good" or "bad" conditioned on the corresponding contexts and the scoring criteria. The instruction manual is presented in Tab.~\ref{tab:scoring-criteria}. This binary classification design aligns with Eq.~\ref{eq:asr-criteria} and the hyper-parameter choices detailed in Sec.~\ref{sec:appendix-hyper-params}. 
With the constraints from $S_{gold}$ part of Eq.~\ref{eq:asr-criteria}, we only get responses of distinctly high-quality or low-quality, which can be validated with binary manual annotations. Regarding the margin part of Eq.~\ref{eq:asr-criteria}, we find that the $S_{victim}$ is much more extreme than bounds constrained by Eq.~\ref{eq:asr-criteria} on our final AdvEval adversarial data, which have an average $S_{victim}$ of 18.1 for $R^{+}$, and 85.4 for $R^{-}$. That extremely low/high value of $S_{victim}$ ensures that $E_{victim}$ makes wrong prediction on our AdvEval data as long as the binary label computed from $S_{gold}$ is correct. Therefore, our human verification process is naturally a binary classification problem instead of a regression problem. The rationale for generating binary good/bad adversarial responses is further supported by various evaluators’ training datasets, which are often constructed as positive and negative instances (binary labels)~\citep{yeh-etal-2021-comprehensive}.

The inter-annotator agreements are found to be 0.77, 0.77, and 0.68 for the dialogue, summarization, and question generation tasks respectively. Since achieving perfect accuracy is exceptionally challenging due to the inherent subjectivity of the task~\citep{amidei-etal-2018-rethinking,zheng2023judging,dubois2023alpacafarm}, this inter-annotator agreements suggest good quality of the annotations. Note that the highest inter-annotator agreement in our study is 77\%, which also shows the challenging nature of achieving complete alignment even among knowledgeable human annotators. Critically, the agreement rate between $E_{gold}$ and human annotators closely mirrors this highest inter-annotator score, indicating the reliability of our gold evaluation as human proxy to guide the optimization process.

\end{document}